# A Decision Theoretic Approach to Targeted Advertising

**David Maxwell Chickering and David Heckerman**
Microsoft Research
Redmond WA, 98052-6399
dmax@microsoft.com
heckerma@microsoft.com

## Abstract

A simple advertising strategy that can be used to help increase sales of a product is to mail out special offers to selected potential customers. Because there is a cost associated with sending each offer, the optimal mailing strategy depends on both the benefit obtained from a purchase and how the offer affects the buying behavior of the customers. In this paper, we describe two methods for partitioning the potential customers into groups, and show how to perform a simple cost-benefit analysis to decide which, if any, of the groups should be targeted. In particular, we consider two decision-tree learning algorithms. The first is an "off the shelf" algorithm used to model the probability that groups of customers will buy the product. The second is a new algorithm that is similar to the first, except that for each group, it explicitly models the probability of purchase under the two mailing scenarios: (1) the mail is sent to members of that group and (2) the mail is not sent to members of that group. Using data from a real-world advertising experiment, we compare the algorithms to each other and to a naive mail-to-all strategy.

## 1  INTRODUCTION

Consider an advertiser who has a large list of potential customers for his product. For a specific real example, we will use Microsoft as the advertiser and a Microsoft Network (MSN) subscription as the product of interest. The potential customers are the people who have registered Windows 95 with Microsoft. Because registering Windows 95 involves filling out a questionnaire, Microsoft has access to lots of useful information about all of the potential MSN subscribers. A typical advertising strategy is to mail out advertisements, perhaps including a special offer for a reduced monthly rate, to a set of potential customers in the hopes that this offer will entice them into signing up for MSN.

Before deciding how to target, the advertiser may be able to perform a preliminary study to determine the effectiveness of the campaign. In particular, the advertiser can choose a small subset of the potential customers and randomly mail the advertisement to half of them. Based on the data collected from the experiment, the advertiser can make good decisions about which members of the remaining population should be targeted.

Perhaps the most obvious approach is to mail all Windows 95 registrants the advertisement for MSN. As described by Hughes (1996), such a mass marketing or *mail-to-all* strategy can often be cost effective. Another strategy that has gained a lot of attention in recent years (e.g. Ling and Li, 1998) is to apply machine-learning techniques to identify those customers who are most likely to subscribe to MSN, and concentrate the campaign on this subset. Assuming that there is a cost to mail the special offer, both strategies may yield negative expected return, and it is unlikely that either strategy will yield the optimal expected return.

In this paper, we describe methods for using experimental data to identify groups of potential customers for which targeting those groups will yield high expected profit for the advertiser. Our approach differs from the machine-learning techniques we identified in the literature by explicitly using expected profit instead of expected response as our objective. In Section 2, we describe how to make the decision whether or not to target a particular group by using a simple cost-benefit analysis with the data collected from the experiment. In Section 3, we describe methods for dividing the population into groups, with the specific goal of maximizing revenue. In Section 4, we present the results from applying our techniques to real-world data. Finally, in Section 5, we conclude with a discus-sion of future direction for this work.

## 2  MAKING THE RIGHT DECISION

In this section, we show how to use the data from an experiment to decide whether or not to send an advertisement to a particular set of potential customers. To understand the problem with the obvious strategies, it is useful to consider how an individual will respond to both receiving and not receiving the advertisement. For any individual, there are only four possible response behaviors he can have. The first behavior, which we call *always-buy*, describes a person who is going to subscribe to MSN, regardless of whether or not he receives the advertisement. The second behavior, which we call *persuadable*, describes a person who will subscribe to MSN if he receives the offer and will not subscribe to MSN if he does not receive the offer. The third behavior, which we call *anti-persuadable*, is the opposite of persuadable: the person will subscribe to MSN if and only if he does *not* receive the offer (perhaps this type of person is offended by the advertisement). Finally, the fourth behavior, which we call *never-buy*, describes a person who is not going to subscribe to MSN, regardless of whether he receives the advertisement.

Assuming that the subscription price exceeds the mailing cost, the optimal strategy is to mail the offer to the persuadable potential customers; for each potential customer that is not persuadable, we lose money by targeting him. If we target an always-buyer, we lose both the cost of the mailing and the difference between the regular subscription price (which the always-buyer was willing to pay) and the (potentially reduced) price that we offer in the advertisement. If we target a never-buyer, we lose the cost of the mailing. The worst is to mail to an anti-persuadable person; in this case, we lose both the cost of the mailing and the regular subscription price.

A potential problem with the mail-to-all strategy is that the advertiser is necessarily mailing to all of the always-buy, never-buy and anti-persuadable customers. The likely-buyer strategy can be problematic as well if a large percent of the people who subscribe are always-buyers.

It is very unlikely that we will ever be able to identify individual response behaviors of potential customers. We can, however, use experimental data to learn about the relative composition of response behaviors within groups of potential customers to easily decide whether or not it is profitable to mail to people in those groups.

Let $N_{Alw}$, $N_{Pers}$, $N_{Anti}$, and $N_{Never}$ denote the number of people in some population with behavior always-buy, persuaded, anti-persuaded, and never-buy, respectively, and let $N$ denote the total number of people in that population. Let $c$ denote the cost of sending out the mailing, let $r_u$ denote the revenue that results from an unsolicited subscription, and let $r_s$ denote the revenue that results from a solicited subscription ($r_u$ minus any discount from the offer). The expected gain from mailing to a person in a population with the given composition is

$$-c + \frac{(N_{Alw} + N_{Pers})}{N} \cdot r_s$$

That is, we pay $c$ to send out the mail; if the person is an always buyer (probability $N_{Alw}/N$) or a persuaded person (probability $N_{Pers}/N$), then he will pay $r_s$. If the person has either of the other two behaviors, he will not pay us anything. Similarly, the expected gain from not mailing is

$$\frac{(N_{Alw} + N_{Anti})}{N} \cdot r_u$$

That is, the always-buyers and the anti-persuaded will pay the unsolicited price $r_u$ if they do not receive the advertisement; the other two types of people will not subscribe.

Given our analysis, the decision of whether or not to mail to a member of the population is easy: send out the advertisement to a person if the expected gain from mailing is larger than the expected gain from not mailing.

$$-c + \frac{(N_{Alw} + N_{Pers})}{N} \cdot r_s > \frac{(N_{Alw} + N_{Anti})}{N} \cdot r_u$$

Or equivalently:

$$\frac{(N_{Alw} + N_{Pers})}{N} \cdot r_s + - \frac{(N_{Alw} + N_{Anti})}{N} \cdot r_u - c > 0 \quad (1)$$

We call the left side of the above inequality the *expected lift in profit*, or *ELP* for short, that results from the mailing.

Both fractions in the above equation are identifiable (that is, they can be estimated from data). In particular, $(N_{Alw} + N_{Pers})/N$ is precisely the fraction of people who will subscribe to MSN if they receive the advertisement, and consequently we can estimate this fraction by mailing to a set of people and keeping track of the fraction of people who sign up for MSN. Similarly, $(N_{Alw} + N_{Anti})/N$ is precisely the fraction of people who subscribe to MSN if they do not receive the advertisement, and consequently we can estimate this fraction by NOT mailing to a set of people and keeping track of the fraction of people who sign up for MSN.

Let $M$ be the binary variable that denotes whether or not a person was sent the mailing, with values $m_0$ (not mailed) and $m_1$ (mailed). Let $S$ be the binary variable that denotes whether or not a person subscribes to MSN, with values $s_0$ (did not subscribe) and $s_1$ (did subscribe). Using these variables, we can re-write the (identifiable) fractions involved in the expected lift as (MLE) probabilities:

$$\frac{(N_{Alw} + N_{Pers})}{N} = p(S = s_1 | M = m_1)$$
$$\frac{(N_{Alw} + N_{Anti})}{N} = p(S = s_1 | M = m_0)$$

Note that MAP estimates for these probabilities can be obtained instead from the given fractions and prior knowledge. Plugging into the definition of ELP (left side of Equation 1) we have:

$$ELP = \quad (2)$$
$$r_s \cdot p(S = s_1 | M = m_1)$$
$$- r_u \cdot p(S = s_1 | M = m_0) - c$$

In the next section, we describe methods for automatically identifying sub-populations that yield large expected lifts in profit as a result of the mailing. As an example, the expected profit from mailing to the entire population (i.e. using the mail-to-all strategy) may be negative, but our methods might discover that there is lots of money to be earned by mailing to the sub-population of females.

## 3 IDENTIFYING PROFITABLE TARGETS

In this section, we describe how to use the data collected from the randomized experiment to build a statistical model that can calculate the ELP for anyone in the population. In particular, we introduce a new decision-tree learning algorithm that can be used to divide the population of potential customers into groups for the purpose of maximizing profit in an advertising campaign.

The experimental data consists of, for each person, a set of values for all distinctions in the domain of interest. The distinctions in the domain necessarily include the two binary variables $M$ (whether or not we mailed to the person) and $S$ (whether or not the person subscribed to MSN) that were introduced in the previous section. We use $\mathbf{X} = \{X_1, ..., X_n\}$ to denote the other distinctions that are in our data. These distinctions are precisely those that we collected in the Windows 95 registration process. The statistical model uses the values for the variables in $\mathbf{X}$ to define the sub-populations that may have different values for ELP.

The statistical model we build is one for the probability distribution $p(S|M, \mathbf{X})$. There are many model classes that can be used to represent this distribution, including generalized linear models, support vector machines, and Bayesian networks. In this paper, we concentrate on decision trees which are described by (e.g.) Breiman, Friedman, Olshen and Stone (1984).

The probability distribution $p(S|M, \mathbf{X})$ can be used to calculate the ELP for anyone in the population. In particular, if we know the values $\{x_1, ..., x_n\}$ for the person, we have:

$$ELP =$$
$$r_s \cdot p(S = s_1 | M = m_1, X_1 = x_1, ..., X_n = x_n)$$
$$- r_u \cdot p(S = s_1 | M = m_0, X_1 = x_1, ..., X_n = x_n)$$
$$- c$$

A *decision tree* $T$ can be used to represent the distribution of interest. The structure of a decision tree is a tree, where each internal node $I$ stores a mapping from the values of a predictor variable $X_j$ (or $M$) to the children of $I$ in the tree. Each leaf node $L$ in the tree stores a probability distribution for the target variable $S$. The probability of the target variable $S$, given a set of values $\{M = m, X_1 = x_1, ..., X_n = x_n\}$ for the predictor variables, is obtained by starting at the root of $T$ and using the internal-node mappings to traverse down the tree to a leaf node. We call the mappings in the internal nodes *splits*. When an internal node $I$ maps values of variable $X_j$ (or $M$) to its children, we say that $X_j$ is the *split variable* of node $I$, and that $I$ *is a split* on $X_j$.

For example, the decision tree shown in Figure 1 stores a probability distribution $p(S|M, X_1, X_2)$. In the example, $X_1$ has two values $\{1, 2\}$, and $X_2$ has three values $\{1, 2, 3\}$. In the figure, the internal nodes are drawn with circles, and the leaf nodes are drawn with boxes. As we traverse down the tree, the splits at each internal node are described by the label of the node and by the labels of the out-going edges. In particular, if the current internal node of the traversal is labeled with $X_i$, we move next to a child of that node by following the edge that is labeled with the given value $x_i$.

Given values $\{X_1 = 1, X_2 = 2, M = m_0\}$ for the predictors, we obtain $p(S|X_1 = 1, X_2 = 2, M = m_0\}$ by traversing the tree in Figure 1 as follows (the traversal for this prediction is emphasized in the figure by dark edges). We start at the root node of the tree, and see that the root node is a split on $X_2$. Because $X_2 = 2$, we traverse down the right-most child of the root. This next internal node is a split on $X_1$, which is has value 1, so we move next to the left child of this node. Finally, because $M = m_0$, we move to the right child, which is a leaf node. We extract the conditional distribution directly from the leaf, and conclude that $p(S = s_1|X_1 = 1, X_2 = 2, M = m_0) = 0.2$ and $p(S = s_0|X_1 = 1, X_2 = 2, M = m_0) = 0.8$.

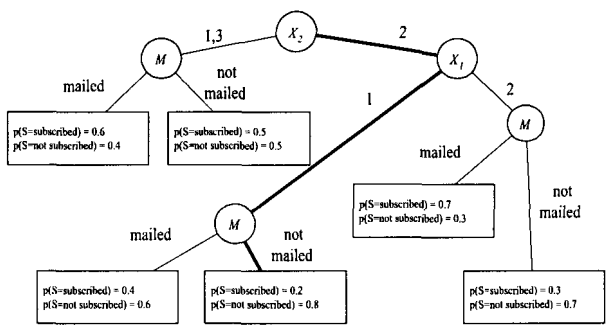

Figure 1: Example decision tree for the distribution $p(S|M, \mathbf{X})$

As an example of how we would use this tree for targeting, suppose we have a potential customer with values $X_1 = 1, X_2 = 2$, and we would like to decide if we should mail to him. Further suppose that $r_u = 10$, $r_s = 8$, and $c = 0.5$. We plug these constants and the probabilities extracted from the tree into Equation 2 to get:

$$ELP = 8 \times 0.4 - 10 \times 0.2 - 0.5 = 0.7$$

Because the expected lift in profit is positive, we would send the mailing to the person. Note that under the given cost/benefit scenario we should not send the mailing to a person for which $X_2 = 1$ or $X_2 = 3$, even though mailing to such a person increases the chances that he will subscribe to MSN.

There are several types of splits that can exist in a decision tree. A *complete* split is a split where each value of the split variable maps to a separate child. Examples of complete splits in the figure are splits on the binary variables. Another type is a *binary* split, where the node maps one of the values of the split variable to one child, and all other values of the split variable to another. The root node is a binary split in the figure.

Decision trees are typically constructed from data using a greedy search algorithm in conjunction with a scoring criterion that evaluates how good the tree is given the data. See Breiman et al. (1984) for examples of such algorithms. Buntine (1993) applies Bayesian scoring to grow decision trees; in our experiments, we use a particular Bayesian scoring function to be described in Section 4. Friedman and Goldszmidt (1996) and Chickering, Heckerman and Meek (1997) both grow decision trees to represent the conditional distributions in Bayesian networks.

The objective of these traditional decision-tree learning algorithms is to identify the tree that best models the distribution of interest, namely $p(S|M, \mathbf{X})$. That is, the scoring criterion evaluates the predictive accuracy of the tree. In our application, however, the primary objective is to maximize profit, and although the objectives are related, the tree that best models the conditional distribution may not be the most useful when making decisions about who to mail in our campaign. We now consider a modification that can be made to a standard decision-tree learning algorithm that more closely approximates our objective.

Recall that the expected lift in profit is the difference between two probabilities: one probability where $M = m_1$ and the other probability where $M = m_0$. Consequently, it might be desirable for the decision-tree algorithm (or any statistical model learning algorithm) to do its best to model the *difference* between these two probabilities rather than to directly model the conditional distribution. In the case of decision trees, one heuristic that can facilitate this goal is to insist that there be a split on $M$ along any path from the root node to a leaf node in the tree.

One approach to ensure this property, which is the approach we took in our experiments, is to insist that the last split on any path is on $M$. Whereas most tree learning algorithms grow trees by replacing leaf nodes with splits, algorithms using this approach need to be modified to replace the last split (on $M$) in the tree with a subtree that contains a split on some variable $X_i \in \mathbf{X}$, followed by a (last) split on $M$ for each child of the node that splits $X_i$.[1] An example of such a replacement is shown in Figure 2. In Figure 2a, we show a decision tree where the last split on every path is on $M$. In Figure 2b we show a replacement of one of these splits that might be considered by a typical learning algorithm.

Note that because leaves used to compute the ELP for any person are necessarily siblings in these trees, it is easy to describe an advertising decision in terms of the other variables. In particular, any path from the root node to an $M$ split describes a unique partition of the population, and the sibling leaf nodes determine the mailing decision for all of the members of that population. As an example, a path in the tree to a split on $M$ might correspond to males who have lots of memory in their computer; the fact that this group of people has a high or low ELP may be particularly interesting to the advertiser.

---

[1]In fact, our implementation of this algorithm does not explicitly apply the last split in the tree. Instead, our scoring criterion is modified to evaluate the tree *as if* there was a last split on $M$ for every leaf node.

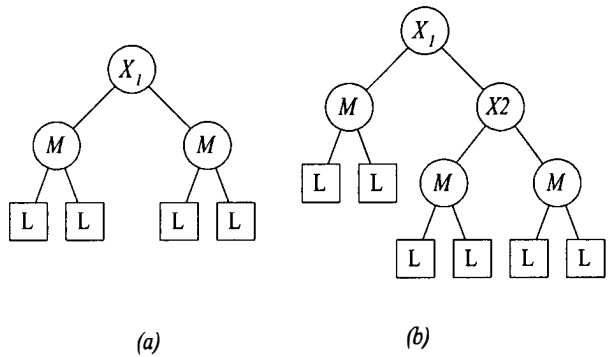

Figure 2: Example of learning trees with the last split required to be on $M$. (a) A candidate solution considered by the learning algorithm and (b) a replacement of the last split to create a new candidate solution.

The hope is that forcing the split on $M$ will steer learning algorithms to trees that are good at predicting ELP. Because we are *forcing* the last split on $M$, however, the final tree may consist of splits on $M$ that do not yield statistically significant differences between the probabilities in sibling leaf nodes. This is potentially problematic, because any such differences are amplified when computing ELP (see Equation 2), and this may lead to bad decisions, particularly in situations when the response benefit is particularly high.

To avoid the problem of statistical insignificance, we post-process the decision trees. In particular, we first remove all of the (last) splits on $M$ in the final tree if doing so increases the score (according to whatever scoring criterion we used to grow the tree). Next, we repeat the following two steps until no change to the tree is made: (1) delete all *last* non-$M$ splits, (2) if any leaf node does not have a parent that is a split on $M$, replace that leaf node with a split on $M$ if doing so increases the score for the model.

In the following section, we evaluate how well a greedy tree-growing algorithm performs using the techniques described in this section.

## 4 EXPERIMENTAL RESULTS

In this section, we present the results from applying two greedy decision-tree learning algorithms to the data collected from an experiment in advertising for MSN subscriptions. The first algorithm, which we call *FORCE*, searches for trees that have the last split on $M$, and then post-processes the tree as described in the previous section. The second algorithm, which we call *NORMAL*, simply tries to maximize the scoring criterion, without forcing any splits on $M$.

The MSN advertising experiment can be described as follows. A random sample of Windows 95 registrants was divided into two groups. People in the first group, consisting of roughly ninety percent of the sample, were mailed an advertisement for an MSN subscription, whereas the people in the other group were not mailed anything. After a specified period of time the experiment ended, and it was recorded whether or not each person in the experiment signed up for MSN within the given time period. The advertisement did not offer a special deal on the subscription rate (that is, $r_s = r_u$).

We evaluated the two algorithms using a sample of approximately 110 thousand records from the experimental data. Each record corresponds to a person in the experiment who has registered Windows 95. For each record, we know whether or not an advertisement was mailed ($M$), and whether or not the person subscribed to MSN within the given time period ($S$). Additionally, each record contains the values for 15 variables; these values were obtained from the registration form. Examples of variables include gender and the amount of memory in the person's computer.

We divided the data into a training set and a testing set, consisting of 70 percent and 30 percent, respectively, of our original sample. Using the training set, we built trees for the distribution $p(S|M, \mathbf{X})$ using the two algorithms FORCE and NORMAL.

For both algorithms, we used a Bayesian scoring criterion to evaluate candidate trees. In particular, we used a uniform parameter prior for all tree parameters, and a structure prior of $0.001^K$, where $K$ is the number of free parameters that the structure can support. Both algorithms were simple greedy searches that repeatedly grew the tree by applying binary splits until reaching a local maximum in the scoring criterion.

To evaluate an algorithm given a cost-benefit scenario (i.e. given $c$ and $r_s = r_u$), we used the test set to estimate the expected revenue per person obtained from using the resulting tree to guide the mailing decisions. In particular, for each record in the test set, we calculated the expected lift in profit using Equation 2, and decided to send the mail if and only if the lift was positive. We should emphasize that the known value for $M$ in each record was ignored when making the decisions; the values for $M$ are "plugged in" directly to Equation 2. Next, we compared our recommendation (mail or do not mail) to what actually happened in the experiment. If our recommendation did not match what happened, we ignored the record and moved on to the next. Otherwise[2], we checked whether or not the corresponding person subscribed to MSN; if he did, we added $r_s - c$ to our total revenue, and if he did not, we added $-c$. Finally, we divided our total revenue by the number of records for which our recommendation matched the random assignment in the experiment to get an expected revenue per person.

For comparison purposes, we also calculated the per-person expected revenue for the simple strategy of mailing to everyone. Then, for both of the algorithms, we measured the *improvement* in the corresponding per-person revenue over the per-person revenue from the mail-to-all strategy. We found that comparing the algorithms using these improvements was very useful for analyzing multiple cost/benefit scenarios; the improvement from using a tree strategy over using the mail-to-all strategy converges to a small number as the benefit from the advertisement grows large, whereas the per-person revenue from a tree strategy will continue to increase with the benefit.

Figure 3 shows our results using a single $c = 42$ cents and varying $r_s = r_u$ from 1 to 15 dollars. For both algorithms, the improvement in the per-person revenue over the mail-to-all per-person revenue is plotted for each value of $r_s$.

The new algorithm FORCE slightly outperforms the simple algorithm NORMAL for benefits less than ten dollars, but for larger benefits the trees yield identical decisions. Although the improvements over the mail-to-all strategy decrease with increasing subscription revenue, they will never be zero, and the strategy resulting from either tree will be preferred to the mail-to-all strategy in this domain. The reason is that both models have identified populations for which the mailing is either independent of the subscription rate, or for which the mailing is actually detrimental.

## 5 DISCUSSION

In this paper we have discussed how to use machine-learning techniques to help in a targeted advertising campaign. We presented a new decision-tree learning algorithm that attempts to identify trees that will be particularly useful for maximizing revenue in such a campaign. Because experimental data of the type used in Section 4 is difficult to obtain, we were only able to evaluate our algorithm in a single domain.

An interesting question is why the new approach only provided marginal improvement over the simpler algorithm. It turns out that for the MSN domain, all of the trees learned using the off-the-shelf algorithm had splits on $M$ for the majority of the paths from the root to the leaves. That is, the condition that motivated our new algorithm in the first place is *almost* satisfied using the simple algorithm. We expect that in general this will not be the case, and that our algorithm will prove to lead searches to better trees for targeting.

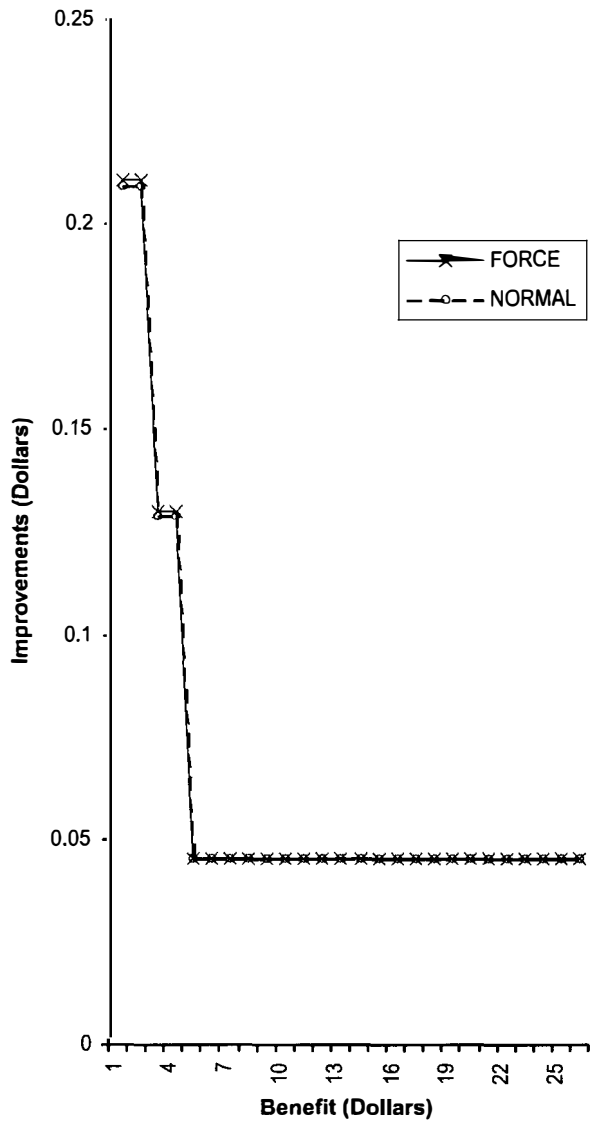

Figure 3: Expected per-person improvement over the mail-to-all strategy for both algorithms, using $c = 42$ cents and varying $r_s = r_u$ from 1 to 15

---

[2]In our experiments, the number of times that our recommendation matched the experiment ranged from a low of roughly 5,000 times to a high of roughly 25,000 times

An obvious alternative approach that meets our split-on-$M$ criterion is to make the *first* split in the tree on $M$. Although our heuristic criterion is met, the approach clearly does not encourage a greedy algorithm to identify trees that predict $ELP = p(S = s_1|M = m_1) - p(S = S_1|M = m_0)$. In fact, this approach is equivalent to independently learning two decision trees: one for the data where $M = m_0$ and another for the data where $M = m_1$. In experiments not presented in this paper, we have found that the approach results in significantly inferior trees.

An interesting extension to this work is to consider campaigns where the advertisement offers a special price. In fact, if the experiment consisted of mailing advertisements of various discounts, we could use our techniques to simultaneously identify the best discount and corresponding mailing strategy.